\documentclass[pdflatex,sn-mathphys-num]{sn-jnl}

\usepackage{graphicx}%
\usepackage{multirow}%
\usepackage{amsmath,amssymb,amsfonts}%
\usepackage{amsthm}%
\usepackage{mathrsfs}%
\usepackage[title]{appendix}%
\usepackage{xcolor}%
\usepackage{textcomp}%
\usepackage{manyfoot}%
\usepackage{booktabs}%
\usepackage{algorithm}%
\usepackage{algorithmicx}%
\usepackage{algpseudocode}%
\usepackage{listings}%

\theoremstyle{thmstyleone}%
\theoremstyle{thmstyletwo}%

\theoremstyle{thmstylethree}%

\begin{document}

\title[Article Title]{Subsampled Randomized Fourier GaLore for Adapting Foundation Models in Depth-Driven Liver Landmark Segmentation}


\author[1]{\fnm{Yun-Chen} \sur{Lin}}

\author[1,2]{\fnm{Jiayuan} \sur{Huang}}

\author[1]{\fnm{Hanyuan} \sur{Zhang}}

\author[1]{\fnm{Sergi} \sur{Kavtaradze}}

\author[1]{\fnm{Matthew} \sur{J. Clarkson}}

\author[1,3]{\fnm{Mobarak I.} \sur{Hoque}}

\affil[1]{\orgdiv{UCL Hawkes Institute and Dept of Medical Physics and Biomedical Engineering}, \orgname{University College London}, \orgaddress{\country{UK}}}
\affil[2]{\orgdiv{Visual Understanding Research Group, Dept of Informatics}, 
\orgname{King's College London}, 
\orgaddress{\country{UK}}}
\affil[3]{\orgdiv{Division of Informatics, Imaging and Data Science}, \orgname{The University of Manchester}, \orgaddress{\country{UK}}}


\abstract{\textbf{Purpose:} Accurate detection and delineation of anatomical structures in medical imaging are critical for computer-assisted interventions, particularly in laparoscopic liver surgery where 2D video streams limit depth perception and complicate landmark localization. While recent works have leveraged monocular depth cues for enhanced landmark detection, challenges remain in fusing RGB and depth features and in efficiently adapting large-scale vision models to surgical domains.

\textbf{Methods:} We propose a depth-guided liver landmark segmentation framework integrating semantic and geometric cues via vision foundation encoders. We employ Segment Anything Model V2 (SAM2) encoder to extract RGB features and Depth Anything V2 (DA2) encoder to extract depth-aware features. To efficiently adapt SAM2, we introduce SRFT-GaLore, a novel low-rank gradient projection method that replaces the computationally expensive SVD with a Subsampled Randomized Fourier Transform (SRFT). This enables efficient fine-tuning of high-dimensional attention layers without sacrificing representational power. A cross-attention fusion module further integrates RGB and depth cues. To assess cross-dataset generalization, we also construct a new Laparoscopic Liver Surgical Dataset (LLSD) as an external validation benchmark.

\textbf{Results:} On the public L3D dataset, our method achieves a 4.85\% improvement in Dice Similarity Coefficient (DSC) and a 11.78-point reduction in Average Symmetric Surface Distance (ASSD) compared to the D2GPLand. To further assess generalization capability, we evaluate our model on LLSD dataset. Our model maintains competitive performance and significantly outperforms SAM-based baselines, demonstrating strong cross-dataset robustness and adaptability to unseen surgical environments.

\textbf{Conclusion:} These results demonstrate that our SRFT-GaLore-enhanced dual-encoder framework enables scalable and precise segmentation under real-time, depth-constrained surgical settings.}





\maketitle

\section{Introduction}\label{sec1}
Minimally invasive surgery (MIS) has been widely adopted across various surgical disciplines for its clear advantages over traditional open surgery, including smaller incisions, reduced risk of infection, and faster postoperative recovery. Laparoscopic liver surgery (LLS), a MIS technique, has become an increasingly preferred and safe approach for treating liver tumors, particularly in oncologic cases. It is indicated for a broad spectrum of hepatic pathologies, including benign tumors (e.g., hemangioma, adenoma), primary malignancies such as hepatocellular carcinoma (HCC), and secondary liver metastases, particularly those of colorectal origin.

Despite its clinical benefits, LLS remains technically demanding. Surgeons rely on monocular laparoscopic cameras that provide limited depth perception, hindering accurate localization of tumors and critical anatomical landmarks. To overcome these challenges, Augmented Reality (AR) has been proposed to overlay preoperative 3D data (e.g., CT or MR) onto intraoperative 2D views \cite{ramalhinho2023value}. Although recent 3D–2D registration methods show promise \cite{koo2022automatic, labrunie2023automatic}, achieving precise alignment remains difficult under real surgical conditions \cite{ali2025objective}, highlighting the need for reliable intraoperative cues such as anatomical landmarks. 

Recent advances like D2GPLand \cite{pei2024depth} leverage monocular depth estimation for landmark detection.However, its reliance on pre-generated depth maps creates a two-stage process, introducing potential latency that limits real-time application. Separately, this approach has also been shown to suffer from unstable training. To address these limitations, we propose a multimodal liver landmark segmentation framework with a cross-attention fusion module for enhanced RGB–depth interaction. The framework integrates state-of-the-art visual encoders, Segment Anything Model V2 (SAM2)~\cite{ravi2024sam} and Depth Anything V2 (DA2)~\cite{yang2024depth}, which are efficiently finetuned using our novel SRFT-GaLore approach. While finetuning such enormous foundation models simultaneously is typically computationally prohibitive, we introduce SRFT-GaLore, a low-rank optimization approach to address this scalability challenge by reducing memory and computational cost.

We evaluated our method on two datasets: the Liver Landmark Segmentation Dataset (LLSD), an annotated version of the LLR dataset~\cite{rabbani2022methodology} that we proposed, and the publicly available L3D dataset~\cite{pei2024depth}. Both datasets contain RGB images with three clinically relevant landmark classes: ridge, falciform ligament, and silhouette, enabling standardized benchmarking and direct comparison with existing approaches such as D2GPLand. Experimental results demonstrate that our model achieves more effective multimodal feature representation and consistently surpasses D2GPLand and other existing models in landmark localization accuracy and training stability, highlighting its potential for real-time, AR-assisted liver surgery.

Our key contributions are summarized as follows: 
\begin{enumerate}
    \item We develop a novel multimodal landmark detection framework featuring a cross-attention fusion module that effectively integrates semantic and geometric features from foundation models SAM2~\cite{ravi2024sam} and DA2~\cite{yang2024depth}, achieving stable training and high localization accuracy.
    \item We enhance scalable fine-tuning by introducing SRFT-GaLore, which replaces the standard SVD in GaLore with a Subsampled Randomized Fourier Transform.
    \item We introduce the Liver Landmark Segmentation Dataset (LLSD), curated and annotated a new unseen test set from a separate domain (LLR dataset~\cite{rabbani2022methodology}) to robustly evaluate cross-dataset generalization.

\end{enumerate}

\section{Background}\label{sec2}
Precise 3D–2D registration is crucial for AR-assisted laparoscopic liver surgery (LLS) but remains a significant challenge \cite{ali2025objective}. Using anatomical landmarks as intraoperative cues is a promising strategy to improve registration robustness. While recent methods like D2GPLand~\cite{pei2024depth} leverage monocular depth for this task, their reliance on a two-stage process, generating depth maps separately before detection introduces potential latency and training instability. To address these limitations, our proposed end-to-end framework integrates recent advances in visual foundation models and parameter-efficient fine-tuning (PEFT).
\\
\textbf{Segment Anything V2 (SAM2): }
Segment Anything V2 builds upon the Hiera architecture~\cite{nawrot2021hierarchical}, a hierarchical vision transformer composed of four convolutional stem layers and 48 MultiScaleBlocks across four stages. The encoder outputs multi-scale feature maps at spatial resolutions of \( H/4 \), \( H/8 \), \( H/16 \), and \( H/32 \), where \( H \) denotes the input height. This pyramid structure facilitates effective extraction of multiscale contextual features, combining high-level semantics with fine-grained spatial details.
 Owing to its strong representational capacity, SAM2 is well suited for dense prediction tasks such as anatomical landmark segmentation, where precise localization and spatial context are essential. In our framework, we employ the SAM2 encoder to leverage its multiscale semantic representations.
 \\
\textbf{Depth Anything V2 (DA2):} 
Depth Anything V2 is a recent foundation model for monocular depth estimation. It is built upon a powerful vision transformer backbone, DINOv2 \cite{oquab2023dinov2}, trained in a self-supervised manner on a large and diverse dataset of 142 million images. In our work, we adapt the DA2 encoder to extract depth-aware features directly from the input images, which complement the visual features for improved landmark detection. 
\\
\textbf{Gradient Low-Rank Projection (GaLore):}
GaLore~\cite{zhao2024galore} is a parameter-efficient fine-tuning (PEFT) method that operates directly on gradients, in contrast to adapter-based methods like LoRA~\cite{hu2022lora}. It projects the full gradient \( G_t \in \mathbb{R}^{m \times n} \) into a low-rank subspace, significantly reducing the memory footprint of optimizer states. Critically, GaLore identifies this subspace by computing the top-$r$ singular vectors of the gradient matrix. This requires performing a computationally expensive Singular Value Decomposition (SVD)
\begin{equation}
    U, \Sigma, V = SVD(G_t), 
\end{equation}
periodically during training. While effective, the complexity of this SVD step (e.g., $ \mathcal{O}(mnr) $) becomes a significant bottleneck when scaling to high-dimensional models like SAM2 and DA2, motivating our work on a more efficient projection strategy.

\section{Methods}\label{sec3}

\begin{figure*}[h]
    \centering
    \includegraphics[width=0.75\linewidth]{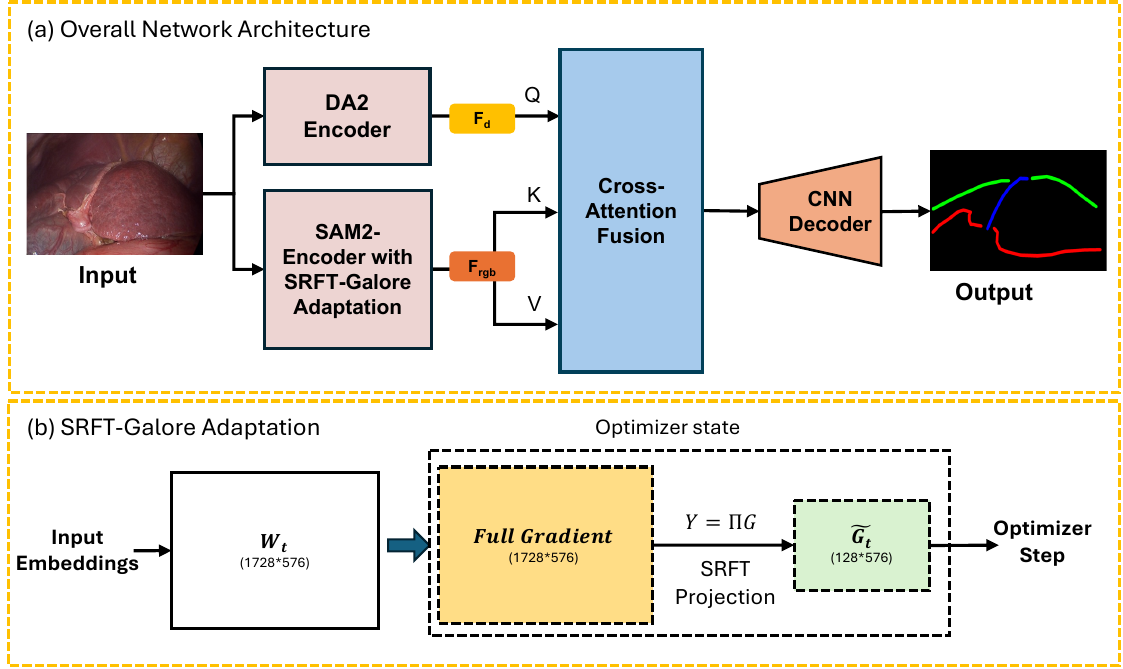}
    \caption{(a) Overview of our proposed model integrating SAM2 and DA2 encoders with cross-attention fusion module. Both encoders are fine-tuned using SRFT-GaLore for efficient adaptation. (b) Illustration of the SRFT-GaLore update, where the full gradient is projected through a SRFT matrix to obtain a compressed low-rank approximation \( \tilde{G}_t \in \mathbb{R}^{128 \times 576} \).}
    \label{fig:model_architecture}
\end{figure*}
Our proposed framework is designed to improve liver landmark segmentation in laparoscopic images by leveraging complementary semantic and geometric representations. The model structure is illustrated in Fig.~\ref{fig:model_architecture}. 

\subsection{Feature Extraction and Adaptation}
We employ a dual-encoder design to capture both semantic and geometric representations from laparoscopic images. The RGB branch adopts SAM2-Large~\cite{ravi2024sam} as a pretrained semantic encoder, providing rich multi-scale contextual features. To adapt SAM2 to the surgical domain, we fine-tune its attention layers using SRFT-GaLore, which enables memory-efficient optimization of high-dimensional parameters by applying low-rank projections to the \texttt{qkv} and \texttt{proj} weights in each attention block.

To complement the semantic features and mitigate appearance ambiguities caused by smoke, specularities, or lighting variations, we incorporate geometric cues through a Depth Anything V2 (DA2) encoder~\cite{yang2024depth}. DA2 extracts depth-aware features from the input RGB images, providing robust geometric context. Similar to SAM2, DA2 is also fine-tuned using SRFT-GaLore to ensure lightweight yet effective adaptation. Together, these two encoders produce complementary feature representations that capture both semantic and geometric aspects of laparoscopic liver scenes.

\subsection{SRFT-GaLore} 
To enable efficient fine-tuning of SAM2 while preserving its representational capacity, we propose SRFT-GaLore: a scalable variant of GaLore. Unlike the original GaLore, which computes projection matrices via truncated SVD with \( \mathcal{O}(mnr) \) complexity, SRFT-GaLore avoids the cost of explicit SVD by using a structured random projection \(\Pi\) to approximate the dominant subspace~\cite{halko2011finding}. SRFT projects \( G_t \) into a randomized low-dimensional subspace that closely approximates its top singular directions. By projecting gradients into this subspace, it retains the most informative components for learning while discarding noisy or redundant variations. This enables efficient adaptation during fine-tuning, reducing memory and computational requirements without compromising convergence or model accuracy.
The matrix \( \Pi \) is defined as:
\begin{equation}
\Pi = S F D,
\label{eq: sfd}
\end{equation}
where \( S \) is a row-sampling matrix selecting \( r \) rows uniformly at random, \( F \) is the unitary Discrete Fourier Transform (DFT) matrix, and \( D\) is a diagonal matrix with random \(\pm1\) entries (sign flipping). This formulation ensures that \( \Pi \) approximately preserves the column space of \( G_t \) while reducing dimensionality~\cite{halko2011finding}. The DFT component \( F \) in \( \Pi \) serves as an orthogonal mixing operator that spreads information uniformly across all coordinates, ensuring that the dominant directions of \( G_t \) are preserved with high probability after subsampling.
We then apply this projection to the gradient matrix:
\begin{equation}
Y = \Pi G_t,
\label{eq:srft_proj}
\end{equation}
which produces a compressed representation that preserves the key directions of \( G_t \). This projection preserves the key characteristics of the original gradient, achieving substantial dimensionality reduction through the mixing and subsampling properties of the SRFT matrix \( \Pi \).
We compute its QR decomposition as \(Y = QR\), provides \( Q\) as the projection matrix \( P_t \), approximating the top-\(r\) singular vectors. This approach allows us to avoid explicit SVD computation while capturing the most informative directions of the gradient for optimization. The final low-rank projected gradient is updated as:
\begin{equation}
    \tilde{G}_t = P_t \cdot \rho_t(P_t^\top G_t),
\end{equation}
enabling memory and computation-efficient optimization. Compared to SVD-based GaLore, SRFT-GaLore reduces projection complexity from \( \mathcal{O}(mnr) \) to \( \mathcal{O}(mn \log n) \), thus enabling scalable fine-tuning for high-dimensional attention layers in SAM2 and DA2 without compromising performance (Fig.~\ref{fig:model_architecture}(b)).

\subsection{Cross-Attention Fusion and Decoding}
To effectively integrate semantic cues from the RGB encoder with geometric cues from the depth encoder, we introduce a cross-attention fusion module that enables depth-guided feature selection. In this module, depth features \(F_d\) act as the query, while RGB features \(F_{rgb}\) serve as the key and value, allowing the network to attend to geometry-aware regions that are semantically relevant. The fused representation is subsequently fed into a lightweight CNN decoder, following the design of D2GPLand~\cite{pei2024depth}, which employs three convolutional stages with skip connections and progressive upsampling for segmentation refinement. The decoder outputs a four-channel prediction corresponding to the background and three anatomical landmark classes. This attention-based fusion pipeline provides more stable training and improved landmark localization compared to convolution-only fusion schemes.

\subsection{Loss Function}
The network is trained end-to-end using a hybrid loss that combines Dice and Binary Cross-Entropy (BCE) terms:
\begin{equation}
\mathcal{L}_{total} = \lambda_1 \mathcal{L}_{Dice} + \lambda_2 \mathcal{L}_{BCE}.
\end{equation}
Here, \(\mathcal{L}_{Dice}\) promotes spatial coherence and boundary accuracy, while \(\mathcal{L}_{BCE}\) improves pixel-wise classification. The loss is computed solely on the final decoder output without auxiliary supervision, with weighting factors \(\lambda_1 = \lambda_2 = 1\).

\subsection{Liver Landmark Segmentation Dataset (LLSD)}
The LLSD dataset was constructed to evaluate cross-dataset generalization in laparoscopic liver landmark segmentation. 
It was derived from the Laparoscopic Liver Resection (LLR) dataset~\cite{rabbani2022methodology}, 
which contains 46 frames of real surgical procedures from 4 patients. Each frame was annotated with multi-class segmentation masks, where each pixel is assigned to one of three landmark classes (anterior ridge, silhouette, and falciform ligament) or background. Compared with L3D, LLSD was annotated with thinner, centerline-like landmark masks in order to emphasize boundary localization. 
This difference in annotation protocol results in slightly lower overlap scores (e.g., DSC, IoU) when models trained on L3D are evaluated on LLSD.


\section{Experiments and Results}\label{sec3}
\subsection{Datasets}
We conducted experiments on two datasets: the publicly available L3D dataset~\cite{pei2024depth} and our newly introduced LLSD. The L3D dataset comprises 1,152 key frames extracted from 39 patients across two medical sites. The dataset is split into 921 training, 122 validation, and 109 test images. To further evaluate the cross-dataset generalization of our model, we created the LLSD dataset. The LLSD dataset originated from a different source than L3D, was used exclusively as an additional, unseen test set. Both datasets are annotated with three types of liver anatomical landmarks: anterior ridge, silhouette, and falciform ligament. The landmark masks provide pixel-level supervision.

\subsection{Implementation Details}
The proposed model was implemented in PyTorch and trained on the L3D dataset~\cite{pei2024depth} and evaluated on both the L3D test set and our LLSD to assess cross-dataset generalization. All images and segmentation masks were resized to \(1024 \times 1024\). The architecture integrates two pretrained encoders, DA2 for capturing depth-aware geometric cues and SAM2 for semantic feature extraction, both fine-tuned using SRFT-GaLore. The projection rank was set to \(r=128\), and projection matrices were refreshed every 50 steps. Data augmentation included random rotations within \(\pm15^\circ\) and horizontal or vertical flips with a probability of 0.5. Training was performed using a 32-bit GaLore-AdamW optimizer with an initial learning rate of \(5 \times 10^{-5}\), cosine annealing decay to \(1 \times 10^{-6}\), and a batch size of 4. The model was trained for 100 epochs, and the checkpoint with the highest validation Dice score was selected for evaluation. All experiments were conducted on a single NVIDIA RTX A6000 GPU.

\subsection{Results}
To evaluate the effectiveness of our model, we compare it with state-of-the-art segmentation methods on the L3D test set and our annotated version of the LLR dataset. These baselines are grouped into two categories: (1) Non-SAM-based approaches, including CNN and transformer models such as UNet~\cite{ronneberger2015u}, COSNet~\cite{labrunie2023automatic}, ResUNet~\cite{xiao2018weighted}, UNet++~\cite{zhou2019unet++}, HRNet~\cite{wang2020deep}, DeepLabV3+~\cite{chen2018encoder}, TransUNet~\cite{chen2021transformers}, and SwinUNet~\cite{cao2022swin}; and (2) SAM-based methods that adapt the Segment Anything Model (SAM), including SAM-Adapter~\cite{wu2025medical}, SAMed~\cite{zhang2023customized}, SAM-LST~\cite{chai2024ladder}, AutoSAM~\cite{hu2023efficiently}, and D2GPLand~\cite{pei2024depth}.

\begin{figure*}[ht]
    \centering
    \includegraphics[width=0.7\linewidth]{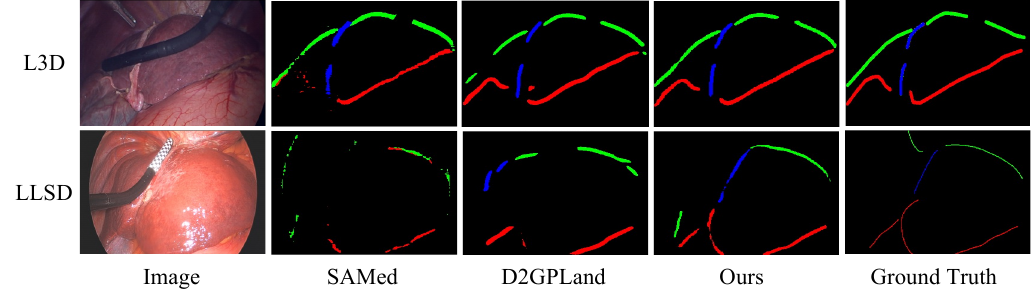}
    \caption{Segmentation performance of ours over SOTA in the classes of ridge, falciform ligament, and silhouette, with the color of red, blue and green.}
    \label{fig:result}
\end{figure*}

Table~\ref{tab: result} summarizes the quantitative results on both the L3D test set and our LLSD. On the L3D test set, our model achieves the best overall performance across all metrics, with a DSC of 68.37\%, IoU of 53.46\%, and ASSD of 47.60. These results represent a 4.85\% improvement in DSC and a 11.78-point reduction in ASSD compared to D2GPLand~\cite{pei2024depth}, and surpass DeepLabV3+~\cite{chen2018encoder} by 8.63\% gain in DSC. To evaluate cross-dataset generalization among the most competitive models, we selected the top-performing SAM-based methods (SAMed and D2GPLand) and our model for evaluation on the LLSD test set. LLSD serves as an external validation set that was not used for training or hyperparameter tuning. Our model maintains superior performance on LLSD, demonstrating robust generalization. We observe a performance shift across all three methods on LLSD, which highlights the challenging domain gap between the datasets. We attribute this shift to differing annotation protocols and variations in average landmark width, as L3D and our LLSD were annotated by different teams.
Fig.~\ref{fig:result} shows qualitative comparisons with SAMed and D2GPLand. Our method yields sharper boundaries and more coherent structures, particularly in complex anatomical regions such as the anterior ridge and falciform ligament. The improvement in ASSD demonstrates our model's surface-level precision, which is crucial for intraoperative guidance and 3D reconstruction tasks.
\begin{table*}[!ht]
\centering
\caption{Comparison with state-of-the-art methods on the L3D test set and LLSD using evaluation metrics of DSC, IoU, and ASSD.}
\label{tab: result}

\scalebox{0.75}{ 
\begin{tabular}{clcccccc} 
\toprule
& & \multicolumn{3}{c}{\textbf{L3D}} & \multicolumn{3}{c}{\textbf{LLSD (External Validation)}} \\ 
\cmidrule(lr){3-5} \cmidrule(lr){6-8} 

\textbf{Category} & \textbf{Model} & \textbf{DSC~$\uparrow$} & \textbf{IoU~$\uparrow$} & \textbf{ASSD~$\downarrow$} & \textbf{DSC~$\uparrow$} & \textbf{IoU~$\uparrow$} & \textbf{ASSD~$\downarrow$}\\
\midrule 

\multirow{8}{*}{\rotatebox{90}{Non-SAM-based}}  & Unet~\cite{ronneberger2015u} & 51.39 & 36.35 & 84.94 & - & - & - \\ 
& COSNet~\cite{labrunie2023automatic}      & 56.24 & 40.98 & 69.22 & - & - & -\\
                               & ResUNet~\cite{xiao2018weighted}     & 55.47 & 40.68 & 70.66  & - & - & -\\
                               & UNet++~\cite{zhou2019unet++}      & 57.09 & 41.92 & 74.31  & - & - & -\\
                               & HRNet~\cite{wang2020deep}       & 58.36 & 43.50  & 70.02  & - & - & -\\
                               & DeepLabv3+~\cite{chen2018encoder}  & 59.74 & 44.92 & 60.86  & - & - & -\\
                               & TransUNet~\cite{chen2021transformers}   & 56.81 & 41.44 & 76.16  & - & - & -\\
                               & SwinUNet~\cite{cao2022swin}    & 57.35 & 42.09 & 72.80   & - & - & -\\
\midrule 
\multirow{5}{*}{\rotatebox{90}{SAM-based}} & SAM-Adapter~\cite{wu2025medical} & 57.57 & 42.88 & 74.31 & - & - & -\\
                               & SAMed~\cite{zhang2023customized}       & 62.03 & 47.17 & 61.55  & 16.63 & 9.6 & 570.87\\
                               & SAM-LST~\cite{chai2024ladder}     & 60.51 & 45.03 & 68.87  & - & - & -\\
                               & AutoSAM~\cite{hu2023efficiently}     & 59.12 & 44.21 & 62.49  & - & - & -\\
                               & D2GPLand~\cite{pei2024depth}    & 63.52  & 48.68  & 59.38  & 22.36 & 13.36 & 287.73\\
\midrule 
\raisebox{-0.5em}{\rotatebox{90}{Ours}} & \textbf{Ours} & \textbf{68.37} & \textbf{53.46} & \textbf{47.60} & \textbf{38.95} & \textbf{24.37} & \textbf{109.11} \\
\botrule
\end{tabular}
}
\end{table*}

\subsection{Ablation Study}

To comprehensively evaluate the design choices in our proposed framework, we conduct ablation studies focusing on: (1) the effect of incorporating depth modality and encoder adaptation; (2) the comparison of different PEFT strategies, including an analysis of projection rank in SRFT-GaLore.
\\
\textbf{Ablation on modality \& encoder tuning:}
We begin by evaluating the impact of depth input and encoder adaptation strategy. As shown in Table~\ref{tab:ablation0}, we compare three configurations: (1) the base model SAMed~\cite{zhang2023customized} using only an fine-tuned SAM encoder without depth input; (2) our proposed model with both SAM2 and DA2 encoders frozen (i.e., no fine-tuning); and (3) our full model in which the SAM2 and DA2 encoder is fine-tuned using SRFT-GaLore. The results demonstrate that incorporating depth input (+Frozen) yields modest improvements over the RGB-only baseline, highlighting the utility of geometric cues. Furthermore, fine-tuning the SAM2 and DA2 encoder with SRFT-GaLore leads to significant gains across all metrics, confirming the effectiveness of task-specific adaptation under memory-efficient constraints.
\\
\begin{table*}[ht] 
\centering
\small
\caption{Ablation study evaluating the effects of excluding depth input and fine-tuning strategy.}
\label{tab:ablation0}
\scalebox{0.8}{
\begin{tabular}{llll}
\toprule
\textbf{Model}  & \textbf{DSC~$\uparrow$} & \textbf{IoU~$\uparrow$} & \textbf{ASSD~$\downarrow$} \\
\midrule
Base & 62.03 & 47.17 & 61.55 \\
+Frozen (DA2+SAM2) & 63.71 & 48.44 & 62.12 \\
+SRFT-GaLore & \textbf{68.37} & \textbf{53.46} & \textbf{47.60} \\
\bottomrule
\end{tabular}
}
\end{table*}

\textbf{Comparison of PEFT Strategies:}
We compare several PEFT methods, including DoRA~\cite{liu2024dora}, MoRA~\cite{jiang2024mora}, AdaLoRA~\cite{zhang2023adalora}, VeLoRA~\cite{miles2024velora}, GaLore~\cite{zhao2024galore}, and SRFT-GaLore. In DoRA, MoRA, AdaLoRA, and VeLoRA, the SAM2 encoder is frozen and only adapters are trained, while GaLore and SRFT-GaLore fine-tune all parameters.
As shown in Table~\ref{tab:ablation1}, SRFT-GaLore achieves the best performance across all metrics, and also yields the fastest training time (634.3s/epoch), despite full-parameter tuning. This highlights the efficiency of combining structured sparsity with low-rank projection.
These findings suggest that SRFT-GaLore not only improves segmentation performance but also offers practical efficiency benefits, making it well-suited for fine-tuning in large-scale medical vision models.
We further ablate the impact of the projection rank \(r\) in SRFT-GaLore, as shown in Table~\ref{tab:ablation2}. A rank of 128 yields the best performance across all metrics, confirming the importance of choosing an appropriate low-rank dimension.

\begin{table*}[!ht]
\centering
\begin{minipage}[t]{0.58\textwidth}
\centering
\caption{Comparison of PEFT strategies on SAM2.}
\scalebox{0.7}{
\begin{tabular}{lcccc}
\toprule
\textbf{PEFT} & \textbf{DSC~$\uparrow$} & \textbf{IoU~$\uparrow$} & \textbf{ASSD~$\downarrow$} & \textbf{Time (s)} \\
\midrule
DoRA~\cite{liu2024dora}        & 66.03 & 50.82 & 55.32 & 1386.7\\ 
MoRA~\cite{jiang2024mora}      & 64.26 & 48.55 & 57.32 & 693.9\\
AdaLoRA~\cite{zhang2023adalora} & 63.03 & 47.49 & 63.03 & 924.9\\
VeLoRA~\cite{miles2024velora}  & 65.98 & 50.76 & 54.32 & 705.6\\
GaLore~\cite{zhao2024galore}   & 66.64 & 51.51 & 51.12 & 668.9\\
\textbf{SRFT-GaLore}           & \textbf{68.37} & \textbf{53.46} & \textbf{47.60} & \textbf{634.3}\\
\bottomrule
\end{tabular}
}

\label{tab:ablation1}
\end{minipage}
\hfill
\begin{minipage}[t]{0.4\textwidth}
\centering
\caption{Ablation study of the impact of rank in SRFT-GaLore.}
\scalebox{0.75}{
\begin{tabular}{lccc}
\toprule
\textbf{Rank} & \textbf{DSC~$\uparrow$} & \textbf{IoU~$\uparrow$} & \textbf{ASSD~$\downarrow$} \\
\midrule
32   & 66.08 & 50.99 & 52.27 \\ 
64   & 66.61 & 51.40 & 49.94 \\
128  & \textbf{68.37} & \textbf{53.46} & \textbf{47.60} \\
256  & 66.26 & 51.19 & 52.31 \\
\bottomrule
\end{tabular}
}
\label{tab:ablation2}
\end{minipage}
\end{table*}

\section{Discussion and Conclusion}\label{sec4}
This paper constructs a liver landmark segmentation dataset (LLSD) and presents a dual-encoder segmentation framework that integrates semantic and geometric cues via cross-attention fusion. By combining a DA2 with a SAM2 encoder which are both fine-tuned with SRFT-GaLore, our model effectively captures complementary representations crucial for accurate liver landmark segmentation. As shown in Table~\ref{tab: result}, our approach achieves the highest Dice score (68.37\%) and the lowest ASSD (47.60), outperforming both SAM-based (D2GPLand) and non-SAM-based (DeepLabv3+) baselines. These improvements stem largely from the proposed SRFT-GaLore fine-tuning strategy, which reduces optimizer state memory by 66.4\% while maintaining accuracy. Ablation studies (Table~\ref{tab:ablation1}) further confirm its effectiveness over other parameter-efficient fine-tuning (PEFT) methods. We attribute SRFT-GaLore’s advantage to its structured random projection, which enhances gradient expressiveness under low-rank constraints and improves boundary delineation without overfitting. To further validate the generalization capability of our framework, we evaluated it on the newly introduced LLSD dataset, an external test set without any retraining or domain adaptation. The model maintained competitive performance and significantly outperformed SAM-based baselines, demonstrating robust cross-dataset generalization to unseen surgical environments. This indicates our method enables better transferability across varied data distributions and imaging conditions.

Despite these gains, the current cross-modal fusion remains shallow and fixed, limiting adaptive integration between semantic and geometric information. Moreover, the CNN-based decoder constrains global context modeling compared to transformer alternatives. Future work will explore deeper cross-modal attention mechanisms, transformer-based decoders, and lightweight architectures for improved global reasoning and real-time deployment. We also plan to investigate the robustness of landmark detection under challenging surgical conditions and extend the framework to support 3D–2D anatomical registration for AR-guided navigation.

\bmhead{Acknowledgements}
This work was supported by the EPSRC under grant [EP/W00805X/1].

\bmhead{Code availability}
The source code of this study is publicly available at: \url{https://github.com/mobarakol/SRFTGaLore_LiverLandmark}.

\bibliography{sn-bibliography}

\end{document}